\documentclass[10pt,twocolumn,letterpaper]{article}

\usepackage{wacv}              % To produce the CAMERA-READY version
\usepackage[accsupp]{axessibility}  % Improves PDF readability for those with disabilities.

\usepackage{amsmath}
\usepackage{enumitem}
\usepackage{mathtools}
\usepackage{amsthm}
\usepackage{multirow}
\usepackage{tikz}
\usepackage{MnSymbol,bbding,pifont}
\usepackage{booktabs}
\usepackage[flushleft]{threeparttable}
\usepackage{algorithm}
\usepackage{algorithmic}
\usepackage{multirow}%,multicol}
\usepackage{wrapfig}
\usepackage[table]{xcolor}

\usepackage{tikz}
\usetikzlibrary{shapes.geometric, arrows.meta, positioning, fit}

\usepackage[most, skins]{tcolorbox}
\newtcolorbox{Summary}{%
  enhanced,
  boxsep=1pt,
  left=2.5pt,
  right=2.5pt,
  top=2.5pt,
  bottom=2.5pt,
  colback=gray!10,
  colframe=black!60
}

\definecolor{wacvblue}{rgb}{0.21,0.49,0.74}
\usepackage[pagebackref,breaklinks,colorlinks,allcolors=wacvblue]{hyperref}

\def\wacvPaperID{1636} % *** Enter the WACV Paper ID here
\def\confName{WACV}
\def\confYear{2026}

\title{RemEdit: Efficient Diffusion Editing with Riemannian Geometry}

\author{
    Eashan Adhikarla \\
    Lehigh University \\
    Bethlehem, Pennsylvania, USA \\
    {\tt\small eaa418@lehigh.edu}
\and
    Brian D. Davison \\
    Lehigh University \\
    Bethlehem, Pennsylvania, USA \\
    {\tt\small bdd3@lehigh.edu}
}

\begin{document}

\makeatletter
\let\@oldmaketitle\@maketitle
\renewcommand{\@maketitle}{\@oldmaketitle
    \begin{center}
    \begin{minipage}{\textwidth}
        \centering
        \vspace{-0.5cm}
        \includegraphics[width=.95\textwidth]{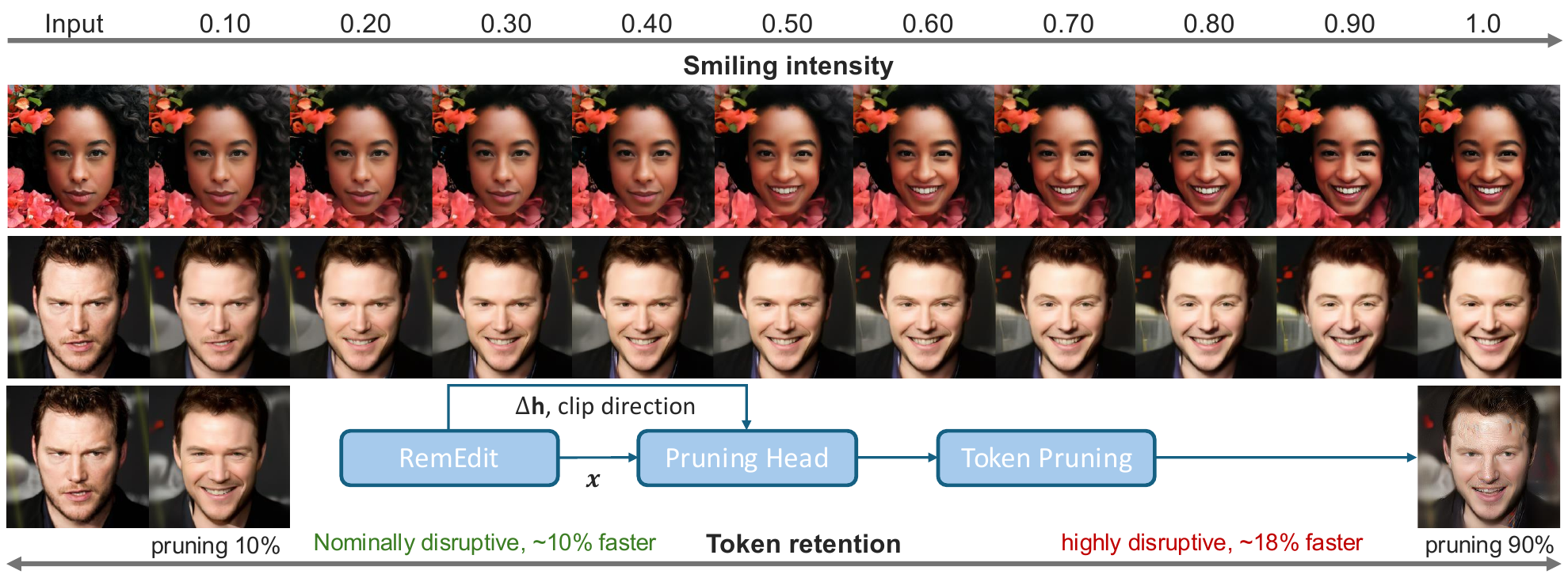}
        \vspace{-0.2cm}
        \captionof{figure}{\label{fig:teaser}RemEdit maintains semantic fidelity under aggressive token pruning; \textbf{90\%} pruning is \textbf{$\sim$18\%} faster yet remains visually acceptable, while \textbf{10\%} pruning is virtually indistinguishable from the unpruned edit and still \textbf{$\sim$10\%} faster.}
    \end{minipage}
    \end{center}
}
\makeatother

\maketitle
\begin{abstract}
    \vspace{-.35in}\\
    Controllable image generation is fundamental to the success of modern generative AI, yet it faces a critical trade-off between semantic fidelity and inference speed. The \textbf{RemEdit} diffusion-based framework addresses this trade-off 
    %, avoiding the compromise between geometric precision and inference speed from which existing methods suffer; RemEdit overcomes this 
    with two synergistic innovations. First, for editing fidelity, we navigate the latent space as a Riemannian manifold. A mamba-based module efficiently learns the manifold's structure,
    %via Christoffel symbols, 
    enabling direct and accurate geodesic path computation for smooth semantic edits. This control is further refined by a dual-SLERP blending technique and a goal-aware prompt enrichment pass from a Vision-Language Model. Second, for additional acceleration, we introduce a novel task-specific attention pruning mechanism. A lightweight pruning head learns to retain tokens essential to the edit, enabling effective optimization without the semantic degradation common in content-agnostic approaches. RemEdit surpasses prior state-of-the-art editing frameworks while maintaining real-time performance under 50\% pruning. Consequently, RemEdit establishes a new benchmark for practical and powerful image editing. Source code: \href{https://github.com/eashanadhikarla/RemEdit}{github.com/eashanadhikarla/RemEdit}.
\end{abstract}

% Controllable image generation is fundamental to the success of modern generative AI, yet it faces a critical trade-off between semantic fidelity and inference speed. The RemEdit diffusion-based framework addresses this trade-off with two synergistic innovations. First, for editing fidelity, we navigate the latent space as a Riemannian manifold. A mamba-based module efficiently learns the manifold's structure, enabling direct and accurate geodesic path computation for smooth semantic edits. This control is further refined by a dual-SLERP blending technique and a goal-aware prompt enrichment pass from a Vision-Language Model. Second, for additional acceleration, we introduce a novel task-specific attention pruning mechanism. A lightweight pruning head learns to retain tokens essential to the edit, enabling effective optimization without the semantic degradation common in content-agnostic approaches. RemEdit surpasses prior state-of-the-art editing frameworks while maintaining real-time performance under 50% pruning. Consequently, RemEdit establishes a new benchmark for practical and powerful image editing. Source code: https://github.com/eashanadhikarla/RemEdit.
\section{Introduction}
\label{sec:intro}

\begin{wrapfigure}[10]{r}{0.18\textwidth}
        \vspace{-3mm}
        \centering
        \includegraphics[width=\linewidth]{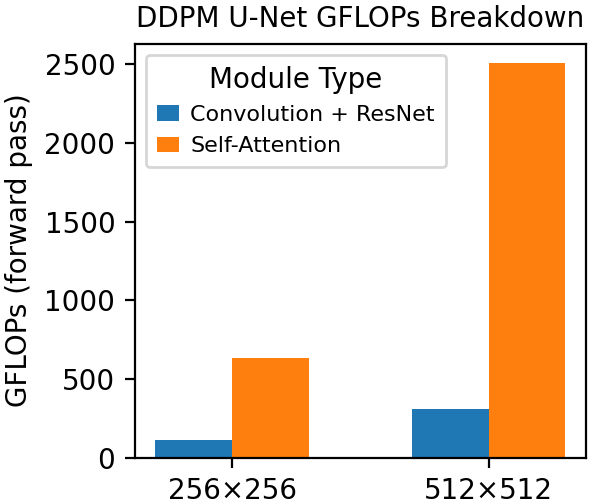}
        \vspace{-0.7cm}\caption{\label{fig:gflops-wrap}GFLOPs breakdown of the DDPM U-Net at 256\textsuperscript{2} and 512\textsuperscript{2}.}
        \vspace{-0.5cm}
    \end{wrapfigure}

    The image synthesis capabilities of diffusion models have shifted the focus of generative AI research towards a more challenging frontier: precise and efficient image editing. While diffusion models~\cite{ho2020denoising,song2020score} have demonstrated SOTA synthesis quality, steering their generative process for precise, user-defined edits remains a challenge. Early approaches focused on text-guided manipulation through costly optimization or fine-tuning~\cite{kim2022diffusionclip}, while others leveraged inversion techniques to enable edits on real images~\cite{Mokady_2023_CVPR}. A pivotal breakthrough was the discovery of the U-Net bottleneck's feature space, termed h-space, which was shown to be remarkably stable and semantically rich~\cite{kwon2022diffusion}. Concurrently, methods like~\cite{hu2023pop,joseph2024iterative} began exploring the internal mechanisms of these models, editing cross-attention maps for more granular control. However, achieving high-fidelity edits that respect the image's original identity often requires intricate tuning. Furthermore, the significant computational cost of the iterative denoising process has motivated a parallel line of research into acceleration, from faster sampling schedules~\cite{ho2020denoising,lu2022dpm} to early efforts in model pruning~\cite{fang2023structural}. This has forced practitioners to choose between powerful but slow methods and faster but less reliable alternatives. The computational bottleneck is particularly severe in the U-Net's self-attention layers, which, as shown in Fig.~\ref{fig:gflops-wrap}, can consume over 80\% of the total GFLOPs in a single forward pass. While this motivates targeting these layers for acceleration, we argue that existing pruning methods are fundamentally misaligned with the goals of high-fidelity image editing.

    To address this trade-off, we propose \emph{RemEdit}, a new framework that jointly optimizes for geometric fidelity and computational efficiency. To our knowledge, \emph{RemEdit} is the first to tackle both challenges for editing within Riemannian diffusion latent spaces. Our work is built on three core pillars of contribution:
    \begin{enumerate}
        \item\textbf{Accurate Geodesic Navigation of $h$-space.} We propose a novel and efficient method for traversing the semantic manifold by learning its local curvature directly. A lightweight Mamba-based predictor efficiently estimates the Christoffel symbols, which in turn define a learnable exponential map. This architecture allows us to solve the geodesic ODE, yielding direct, smooth, and geometrically faithful edits.

        \item\textbf{High-Fidelity Semantic Control.} We achieve superior control and blending through two key innovations. A dual-SLERP mechanism provides principled, artifact-free interpolation within the manifold-aware latent space. This is guided by a goal-aware prompt enrichment strategy that uses a single pass of a Vision-Language Model (Qwen2-VL) to generate a nuanced, context-aware semantic direction.

        \item\textbf{Task-Specific Acceleration.} We introduce a novel task-aware attention pruning method that dramatically accelerates inference without sacrificing edit quality. A lightweight ``PruningHead,'' conditioned on the semantic goal of the edit, learns to preserve the most relevant tokens. This enables aggressive acceleration of irrelevant regions while maintaining integrity of the manipulation.
    \end{enumerate}
    The synergy of these contributions results in a framework that is both more powerful and significantly more practical than prior work, closing the gap between theoretical robustness and real-world usability. More discussions on key challenges motivating this work are in \textbf{Appx.~A}.
    \begin{figure*}[ht]
        \centering
        \includegraphics[width=0.95\linewidth]{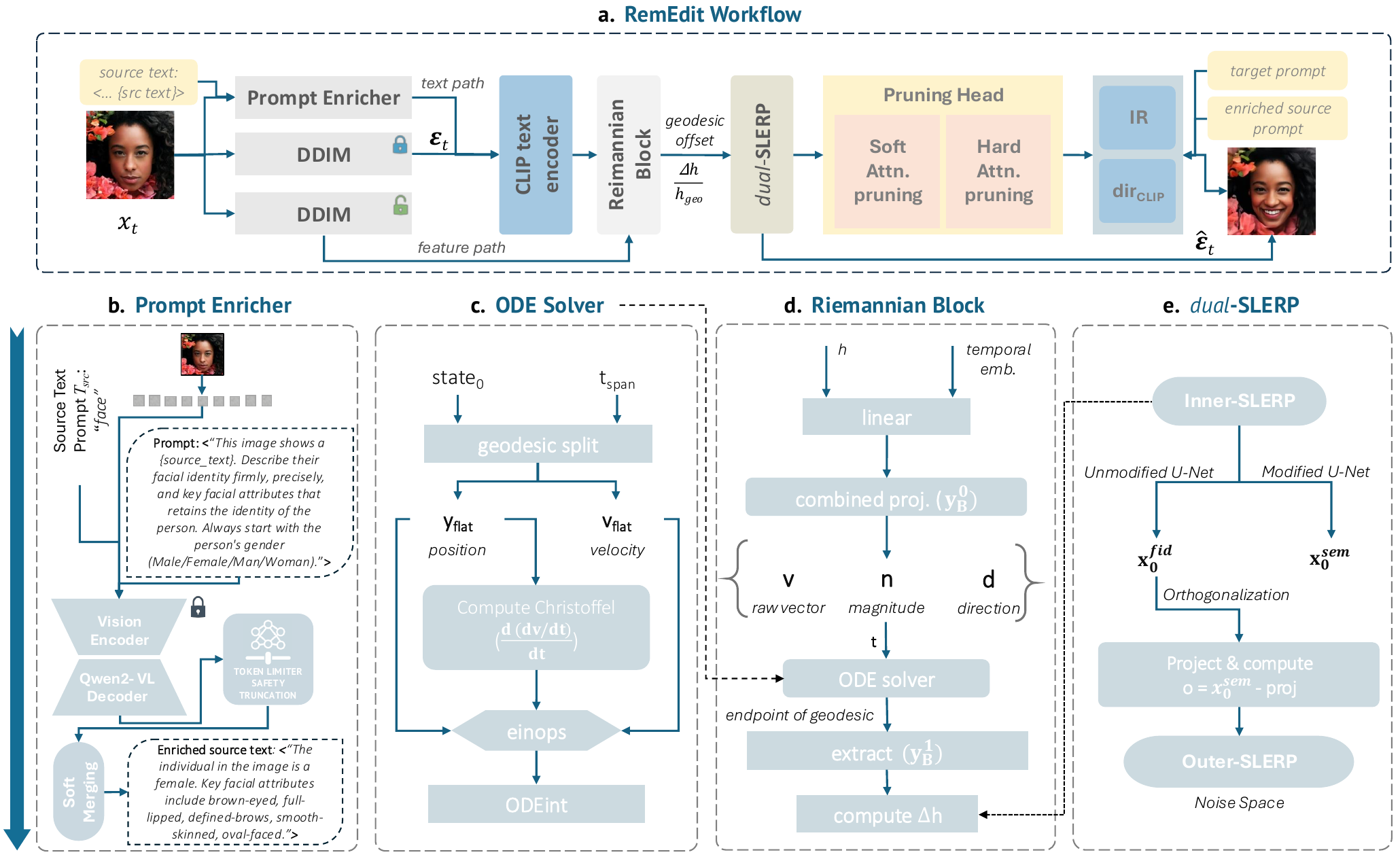}
        \caption{\label{fig:remedit-arch}\textbf{RemEdit Architecture Diagram.} Overview of our diffusion editing pipeline integrating exponential map for geodesic navigation, dual-SLERP interpolation for fidelity control, Qwen2-VL for prompt enrichment. The individual modules flow from top to bottom.}
    \end{figure*}
\section{Related Work}
\label{sec:rw}

% \subsection{Semantic latent structures in $h$-space}
\subsection{Semantic latent structures in \texorpdfstring{$h$}{h}-space}
    % Kwon et al.\ introduced the notion of h-space—the bottleneck features in the U-Net denoiser–-and showed that simple arithmetic in this space can yield consistent semantic edits without retraining \cite{kwon2022diffusion}. This space, denoted as $h_t$ at timestep $t$, serves as a powerful control substrate where semantic edits can be applied via a simple offset, $\Delta{h}$. This discovery, further explored in $h$-edit~\cite{nguyen2025hedit}, catalyzed an extensive line of research as summarized in \cite{park2023understanding}, aimed at decoding and leveraging this latent structure.

    % Subsequent efforts demonstrated the compositional power of $h$-space. Jeong et al.'s  InjectFusion \cite{jeong2024training} enabled training-free image blending by interpolating $h$-space features. PCA-based direction discovery \cite{haas2024discovering,zhang2023unsupervised} revealed interpretable semantic axes like age or gender. Semantic-Diffusion~\cite{haas2024discovering} further applied this space for 3D-consistent generation, and \cite{ijishakin2024h} proposed an inversion strategy within h-space that improves style fidelity during editing \cite{ijishakin2024h}.

    Kwon et al.\ first formalized the U-Net bottleneck ($h$-space) and showed that simple offsets enable reliable edits without retraining~\cite{kwon2022diffusion}. Follow-ups explored compositionality (InjectFusion)~\cite{jeong2024training}, PCA directions/semantic axes~\cite{haas2024discovering,zhang2023unsupervised}, 3D consistency~\cite{haas2024discovering}, and $h$-space inversion for style fidelity~\cite{ijishakin2024h}. Surveys summarize this trend toward interpretable h-space control~\cite{park2023understanding}. We build on this line, but argue that the stability often attributed to linearity is better explained by an underlying Riemannian structure of diffusion dynamics.

    Recent works \cite{park2025navigating,yang2024unleashing} emphasize that $h$-space is more than a transient encoding, but is instead a structured, semantically aligned representation. Our work builds on this, proposing that its stability and linearity emerge from the Riemannian geometry intrinsic to diffusion dynamics.

\subsection{Geodesic and Manifold-Aware Editing}
    % The idea that high-dimensional data lies on a lower-dimensional manifold \cite{fefferman2016testing} has long influenced generative modeling, from early work in Isomap and LLE \cite{tenenbaum1997mapping,doi:10.1126/science.290.5500.2323} to its application in GANs \cite{harkonen2020ganspace,shen2020interfacegan}. While $h$-space has emerged as a reliable latent representation, most editing methods treat it as Euclidean. A growing body of research explores explicit manifold modeling in diffusion. Some methods aimed at meaningful and decodable representations through autoencoders~\cite{preechakul2022diffusion,lu2024hierarchical}. Score-based approaches define post-hoc Riemannian metrics via score function Jacobians \cite{azeglio2025s,saito2025image,park2022riemannian} or minimize path lengths to compute geodesics \cite{park2023understanding}, often incurring substantial computation. Others employ pullback metrics to analyze local geometry \cite{park2023understanding}.

    % In contrast, our method directly learns the manifold connection within $h$-space using a mamba-based exponential map that exponentiates each tangent perturbation into a geodesic step, keeping edits on-manifold while retaining the mamba block’s $O(N)$ efficiency. This enables us to efficiently solve the geodesic ODE without requiring Jacobian estimation, offering a scalable and principled route to geometry-aware semantic editing.

    Manifold assumptions pervade generative modeling, from classic Isomap~\cite{tenenbaum1997mapping}/LLE~\cite{doi:10.1126/science.290.5500.2323} to GAN latent geometry~\cite{fefferman2016testing,harkonen2020ganspace,tenenbaum1997mapping,shen2020interfacegan,doi:10.1126/science.290.5500.2323}. Recent works adapt this view to diffusion via autoencoding latents~\cite{preechakul2022diffusion,lu2024hierarchical}, score Jacobian metrics~\cite{azeglio2025s,saito2025image,park2022riemannian} and shortest paths~\cite{park2023understanding} (with extended discussion in \textbf{Appx. A})%Suppl.\ Sec.~\ref{supp:extended_problem_statement}). 
    In contrast, we learn the $h$-space connection and integrate an exponential map (mamba-based) to take geodesic steps efficiently, avoiding Jacobian estimation while keeping \texttt{O(N)} complexity.

\subsection{Semantic Guidance and Control}

    Training-free T2I editors localize changes either through attention control or implicit masking. Prompt-to-Prompt (P2P) constrains word-level changes by reusing cross-attention maps~\cite{hertz2022prompt} and MasaCtrl~\cite{cao2023masactrl}; LEDITS++~\cite{brack2024ledits++} combines fast inversion with implicit attention masking. % for precise, localized edits. 
    We include both under identical prompts, inversion depths, and seeds in our comparisons and observe that RemEdit achieves stronger locality and identity preservation at comparable or faster runtime. These methods excel when text alone suffices, but are sensitive to phrasing and lack an explicit geometric constraint. Our approach complements them by constraining edits along learned $h$-space geodesics with dual-SLERP to balance edit strength and identity. Zero-shot editors built atop modern T2I pipelines further reduce overhead: Null-Text Inversion (optimization-based inversion)~\cite{Mokady_2023_CVPR}, Negative-Prompt Inversion (one-shot inversion)~\cite{miyake2025negative}, Inversion-Free Editing (InfEdit) with modified sampler; no explicit inversion~\cite{xu2023inversion}, and Lightning-Fast inversion via guided Newton steps~\cite{samuel2023lightning}. These works achieve strong identity retention and speed but remain prompt-dependent. RemEdit is complementary: by operating in unconditional $h$-space with a learned connection and dual-SLERP, it enforces identity and locality by design under the same prompts.

\subsection{Acceleration of Diffusion Models}

    Sampling-level accelerators (DDIM/ DPM-Solver/ distillation) reduce steps~\cite{ho2020denoising,lu2022dpm, salimans2022progressive}, while token-level methods prune attention with training-free (SiTo~\cite{zhang2025training}/AT-EDM~\cite{wang2024attention}), hybrid (DaTo~\cite{zhang2024token}/CAT~\cite{cheng2025cat}), or merging (ToFu~\cite{kim2024token}) strategies. Training-based approaches (DyDiT~\cite{zhao2024dynamic}/LD-Pruner~\cite{castells2024ld}) restructure modules. Other works like EffDiff~\cite{starodubcev2023towards} focus on real-time manipulation. Editing imposes stricter locality than generation; content-agnostic pruning can harm identity in unedited regions. Despite their ingenuity, the aforementioned methods are designed and evaluated primarily for T2I. The task of semantic editing imposes a much stricter constraint: semantic consistency must be preserved across all unedited regions of the image. A content-agnostic pruner, by its nature, cannot distinguish between tokens that are globally redundant (e.g., a patch of blue sky) and tokens that are locally critical for maintaining the identity of an unedited object. Pruning the latter can lead to unacceptable degradation of the source content. We introduce a task-aware pruner tailored to edit locality. %(Extended review in Supplement Sec. A.)
\section{Method}
\label{sec:method}

\subsection{Riemannian Edit Framework}
\label{sec:remedit}

Our complete architecture is illustrated in Fig.~\ref{fig:remedit-arch}.

% \paragraph{Geodesic Navigation via the Exponential Map.}
    A central challenge in semantic editing is that linear manipulations in latent space~\cite{kwon2022diffusion} often fail to preserve image realism. H-space methods typically apply a learned linear offset: $h' = h + \Delta h_{\text{linear}}$. Our results show that geodesic updates improve directional alignment and segmentation consistency with only a small increase in computation, later recovered through pruning. Adding a vector to a latent code can push it off the manifold of natural images\footnote{Linear offsets often leave the manifold, producing edits that no longer preserve the subject’s identity.}, resulting in artifacts. This raises a critical question:
    %
    % \begin{center}
    % \textbf{How can we edit semantically while respecting the data manifold, without high computational cost?}
    % \end{center}
    \begin{Summary}
        \textit{How can we edit semantically while respecting the data manifold, without high computational cost?}
    \end{Summary}
    While prior methods acknowledge this geometric structure, they typically rely on post-hoc computations such as estimating metrics from score function Jacobians or solving expensive optimization problems~\cite{saito2025image,park2022riemannian}.
    
    We estimate the full metric through an ODE solver to capture geodesics curvature and endpoints, while directly learning the manifold’s connection.\footnote{the rules that govern how a vector is transported along a curve.} This allows us to compute geodesics—the straightest possible paths on a curved surface—using the mathematical tool of the exponential map.
    
    In Riemannian geometry, the exponential map, denoted $exp_{p}(v)$, takes a point $p$ on a manifold and a tangent vector $v$ at that point, and maps it to a new point on the manifold by traveling along the geodesic starting at $p$ in the direction $v$ for unit time. The path of this geodesic, $\gamma(t)$, is governed by the geodesic equation, which depends on the manifold's Christoffel symbols $\Gamma^k_{ij}$:
    \begin{equation}\label{eq:geodesic}
        \frac{d^2 \gamma^k}{dt^2} + \boldsymbol{\Gamma}^k_{ij} \frac{d \gamma^i}{dt} \frac{d \gamma^j}{dt} = 0
    \end{equation}
    The key insight is that by learning the Christoffel symbols, we can solve this second-order ODE to efficiently compute the exponential map. As illustrated in our architecture (see Appendix, Fig.~\ref{fig:expomap_arch}), our Riemannian block implements this as follows:
    \textbf{(a)} \textbf{Initialization:} We form an initial point $y_{0}$ on the manifold by combining the input $h$-space feature $h$ with the temporal embedding. \textbf{(b)} \textbf{Tangent Vector Prediction:} From $y_{0}$, we predict an initial tangent velocity vector $v_{0}$ using a linear layer followed by a smooth, norm-preserving tanh-based retraction. \textbf{(c)} \textbf{ODE Integration:} We learn the Christoffel symbols $\Gamma$ using a lightweight Mamba network. We then solve the geodesic equation (Eq.~\ref{eq:geodesic}), formulated as a first-order system, from $t=0$ to $t=1$ using a high-precision adaptive ODE solver. \textbf{(d)} \textbf{Offset Extraction:} The solution to the ODE at $t=1$ gives the endpoint of the geodesic, $\gamma(1)=exp_{h}(v_0)$. We then define our geometrically-sound edit vector as the displacement along this path:
    \begin{equation}
        \Delta h = \exp_h(v_0) - h
    \end{equation}
    This allows us to formulate the $h$-space update in a principled, on-manifold fashion, where the final edited feature is the geodesic endpoint itself:
    \begin{equation}
        h'_{\text{RemEdit}} = h + \Delta h{_\text{geo}} = \exp_h(v_0)
    \end{equation}
    A breakdown of the exponential-map module and ODE solver configuration is included in the \textbf{Appx.~C}.

\subsection{High-Fidelity Control with Dual-SLERP}
\begin{figure}
        \centering
        \includegraphics[width=\linewidth]{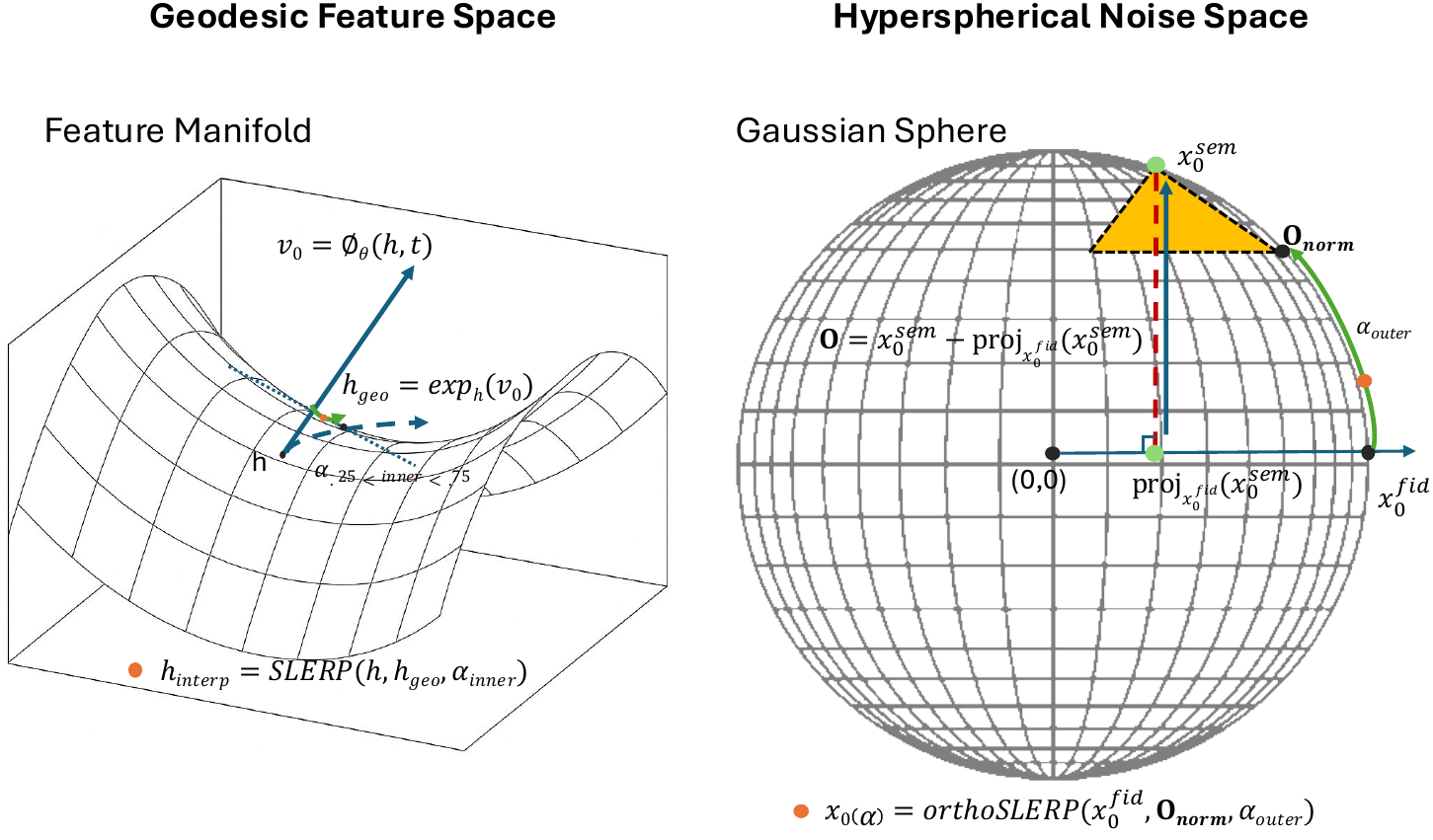}
        \caption{\label{fig:dual-slerp}Dual‐SLERP two‐stage interpolation: \textbf{[Left]} Inner SLERP on the Riemannian feature manifold, blending the original feature $h$ and the geodesically shifted feature $\exp_{h}(v_{0})$ via the interpolation parameter $\alpha_{\mathrm{inner}}$. \textbf{[Right]} Outer orthogonal SLERP on the hyperspherical noise latent space, projecting the semantic prediction $x_{0}^{\mathrm{sem}}$ onto the fidelity latent $x_{0}^{\mathrm{fid}}$, extracting the orthogonal component $o$, and interpolating between $x_{0}^{\mathrm{fid}}$ and $o$ using $\alpha_{\mathrm{outer}}$, thus disentangling attribute edits from identity preservation.}
        \vspace{-.3cm}
    \end{figure}

    The geodesic offset $\Delta{h}$ (Sec.~\ref{sec:remedit}) provides a robust, geometry-aware direction for editing. However, applying it naively as $h' = h + \Delta{h}$ offers no control over the edit's intensity and risks overpowering the original image's identity. This raises the next critical question:
    %
    % {\bf \begin{center}
    %     How can we scale the geodesic offset $\Delta{h}$ to control edit strength without introducing artifacts?
    % \end{center}}
    %
    \begin{Summary}
    % \begin{Summary}[title=\textbf{Research Question 2.}]{}{}
        \textit{How can we scale the geodesic offset $\Delta{h}$ to control edit strength without introducing artifacts?}
    \end{Summary}
    \noindent We hypothesized that the latent features in diffusion models approximate a hyperspherical gaussian distribution. Therefore, interpolation should be performed spherically to remain on the data manifold and preserve statistical properties. Linear interpolation (Lerp) fails to preserve the norm of latent vectors and can ``fall off'' this hypersphere, leading to a collapse in quality. Spherical Linear Interpolation (SLERP)~\cite{shoemake1985animating,song2020denoising}, which travels along the great-circle arc between two points, is the natural choice for this space. We employ this insight in a novel dual-SLERP mechanism, illustrated in Fig.~\ref{fig:dual-slerp}, for two distinct levels of control.

    \vspace{-.1in}
    \paragraph{Inner SLERP (Modulating Edit Strength).}The first stage of our blending strategy occurs within the feature space, as depicted on the left in Fig.~\ref{fig:dual-slerp}. After computing the full geodesic offset $\Delta{h}$ using the Riemannian framework (Sec.~\ref{sec:remedit}), we SLERP between the original feature map h and the fully edited feature map $h_{geo}=h+\Delta{h}$. This interpolation is defined as: $h' = SLERP(h,h_{geo},\alpha_{inner})$. The scalar $\alpha_{inner} \in [0,1]$ acts as a precise knob\footnote{decoupling edit's direction from its magnitude (controlled by SLERP)} to control the strength of the semantic edit in feature space, before it is passed to subsequent layers of the U-Net.

    \vspace{-.1in}
    \paragraph{Outer SLERP (Preserving Global Fidelity).}The second stage occurs at the end of a denoising step to form the final prediction of the clean image, $x_0$. Our framework naturally produces two predictions: a fidelity-preserving latent vector, $x_{0}^{fid}$, from the original U-Net path, and a semantically edited one, $x_{0}^{sem}$, from our edited path. To robustly fuse these while preserving the original image's identity, we use an orthogonalized SLERP in noise space (shown in Fig.~\ref{fig:dual-slerp}). We first project the semantic latent onto the fidelity latent to find the component of the edit that is orthogonal to the original identity: $o=x_{0}^{sem} - proj_{x_{0}^{fid}}(x_{0}^{sem})$. We then interpolate between the fidelity prediction and orthogonal component $o$:
    \begin{equation}
        x_0(\alpha_{outer}) = \Psi(x_{0}^{fid},o,\alpha_{outer})
    \end{equation}
    A broader conceptual comparison of interpolation approaches is in \textbf{Appx.~E}.

\subsection{Goal-Aware Prompt Enrichment}
\label{sec:qwen}

\begin{figure*}
    \centering
    \includegraphics[width=0.9\linewidth]{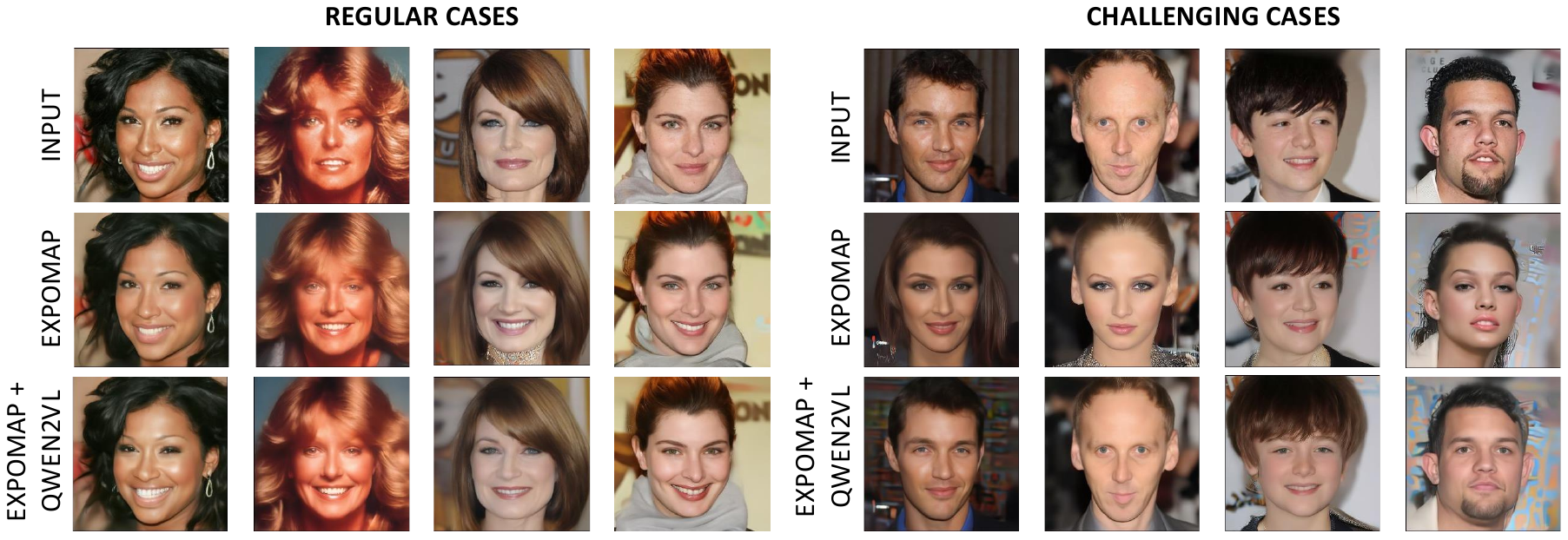}
    \caption{\label{fig:qwen2}Showing through a ``\texttt{makeup}'' task how the usage of Qwen2 VL for fine grained text injection corrects some of the failure cases.}
    \vspace{-0.2cm}
\end{figure*}

    A core challenge faced by prior methods is in text-guided editing ambiguity, as concise prompts often lack the specificity to prevent unrelated attribute changes. For an attribute like makeup, a simple source-target pair such as ``face'' $\rightarrow$ ``face with makeup'' provides insufficient context. The model, seeking to satisfy this minimal constraint, is free to explore neighboring manifolds in the data distribution that may introduce unintended changes, such as altering the perceived gender of the subject. This necessitates providing richer, instance-specific context to the edit direction without adding significant computational overhead and w/o re-labeling text for image pair. We call this a ``no free lunch'' problem, which we solve by introducing a lightweight, single-pass prompt enrichment stage using a pretrained Vision-Language Model (Qwen2-VL). Instead of using a generic source text for all images, we first generate a detailed caption for the specific source image $x_0$ (details shown in Appendix Fig.~\ref{fig:qwen-diagram}). This approach effectively grounds the edit (shown in Fig.~\ref{fig:qwen2}) in the specific visual context of the input image, mitigating unwanted semantic shifts by narrowing the model's exploratory freedom. Further details are in \textbf{Appx.~D}.

\subsection{Task-Specific Attention Pruning}

    Our framework thus far achieves high-fidelity, controllable editing. However, like all modern diffusion models, it remains computationally expensive. %Analysis of the U-Net architecture reveals that t
    The self-attention blocks are the primary bottleneck, consuming substantial resources: over 4$\times$ the GFLOPs of convolution and ResNet blocks at $256\times256$ resolution, and over $8\times$ at $512\times512$ resolution. This makes them ideal for optimization.

    While token pruning has emerged as a viable acceleration strategy, existing methods are designed for general image generation. This raises a critical question:% for our use case: %\textit{\textbf{How can we prune computationally expensive tokens without violating the core constraint of image editing: preserving the semantic content of unedited regions?}} 
    \begin{Summary}
        \textit{How can we prune computationally expensive tokens without violating the core constraint of image editing: preserving the semantic content of unedited regions?}
    \end{Summary}
    For pure generation, minor inconsistencies or artifacts may be acceptable. For editing, they represent failure. 
    If editing a ``smile'' onto a face alters the subject's hair or the background, the edit has failed. This implies that a pruner for editing cannot be content-agnostic; it must be explicitly aware of the editing task. A successful pruning strategy for semantic editing must be conditioned on the edit vector itself. By providing the pruning mechanism with information about the desired semantic change, it can learn to distinguish between tokens that are globally redundant and tokens that are critical for preserving the identity of unedited regions, even if those regions are low-variance.

    We introduce a novel \textbf{task-aware attention pruning} mechanism. Instead of relying on static heuristics, we train a lightweight neural network to dynamically predict token importance, conditioned on both the input features and the semantic goal of the edit.

    The training objective is a weighted sum of two competing losses, designed to balance output quality with computational savings: $\mathbf{L} = \mathcal{L}_{\text{fidelity}} + \lambda_{\text{sparsity}}.\mathcal{L}_{\text{sparsity}}$. First, a Fidelity Loss teaches the pruner what to keep by forcing its output to match the original, unpruned model's output. Second, a Sparsity Loss encourages the model to be efficient by pushing it to prune as many tokens as possible. During inference, this trained pruner $\mathcal{P}_{\theta}$ performs non-differentiable hard pruning.

    Given input features $X\in\mathbb{R}^{B \times C \times H \times W}$ and a task-defining semantic vector $\mathbf{d}_{\text{edit}} \in \mathbb{R}^{D_{\text{clip}}}$, we reshape $X$ into a token sequence $T \in \mathbb{R}^{B \times N \times C}$, where $N = H \cdot W$. A learned pruning function $\mathcal{P}_\theta$ estimates token importance:
    \[
        S = \mathcal{P}_\theta(T, \mathbf{d}_{\text{edit}}) \in [0,1]^{B \times N}
    \]
    Top-$k$ indices are selected with $k = \lfloor N \cdot (1 - \rho) \rfloor$, and attention is computed only over retained tokens:
    \[
        A = \text{Softmax}\left(\frac{Q_{\text{kept}} K_{\text{kept}}^\top}{\sqrt{C}}\right) V_{\text{kept}}
    \]
    The attention output is restored to the full token space and projected, yielding the final output $X_{\text{out}}$:
    \[
        X_{\text{out}} = X + W_{o}(\text{Scatter}(A, \mathcal{I}_{\text{keep}}))
    \]
    See \textbf{Appx.~B} for more detail on pruning. 
    %We provide more mathematical details on pruning in Suppl. Sec.~\ref{supp:extended_details_prune}.
\begin{figure}
    \centering
    \includegraphics[width=0.95\linewidth]{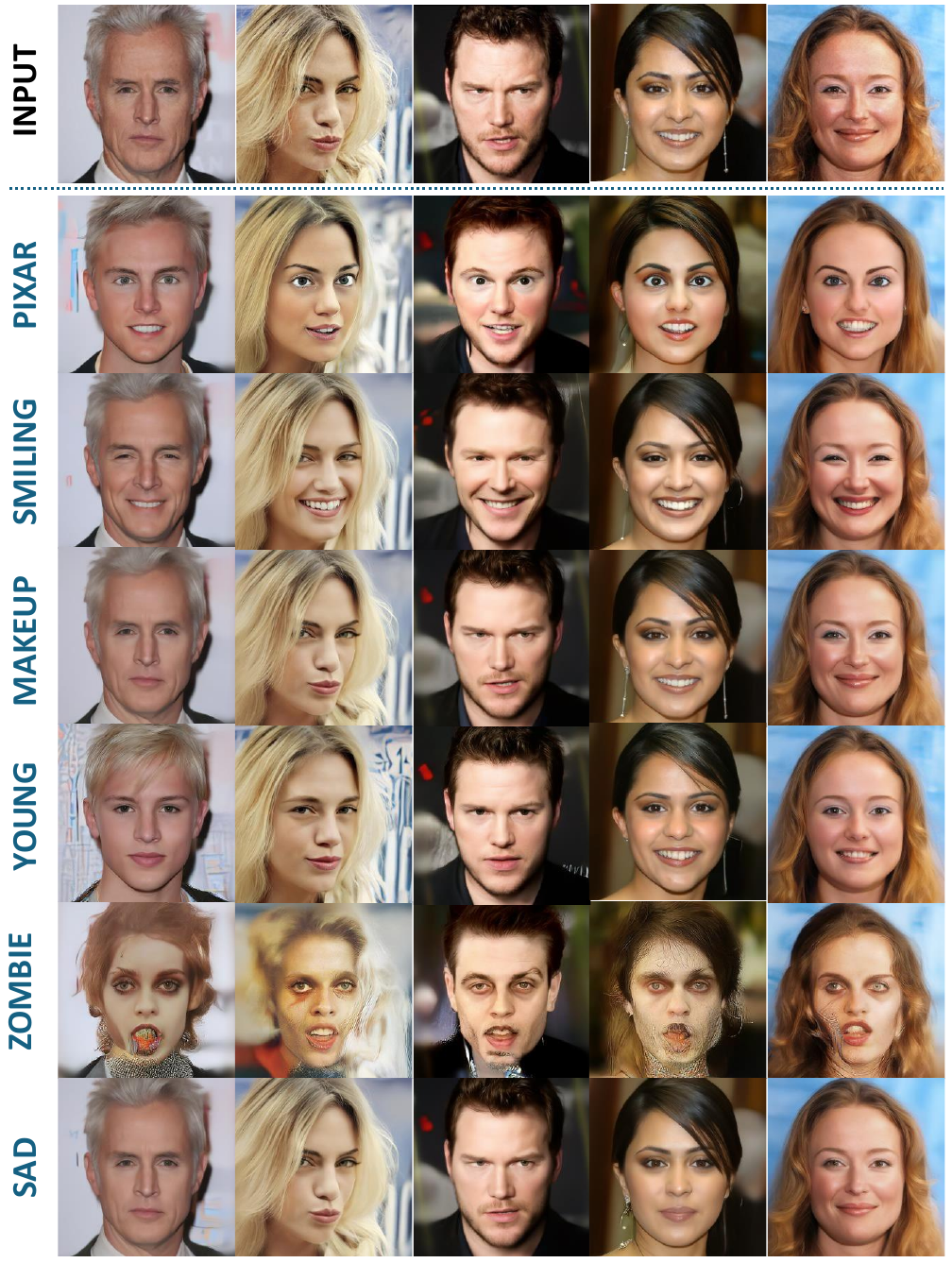}
    \caption{Editing results of RemEdit on CelebA-HQ dataset where attributes \{``\texttt{Sad}'', ``\texttt{Smiling}'', ``\texttt{Makeup}'', ``\texttt{Young}''\} are Human in-distribution and \{``\texttt{Zombie}'', ``\texttt{Pixar}''\} are Human out-of-distribution.}
    \label{fig:various_attr}
    \vspace{-0.5cm}
\end{figure}
\begin{figure}[t]
    \centering
    \includegraphics[width=\linewidth]{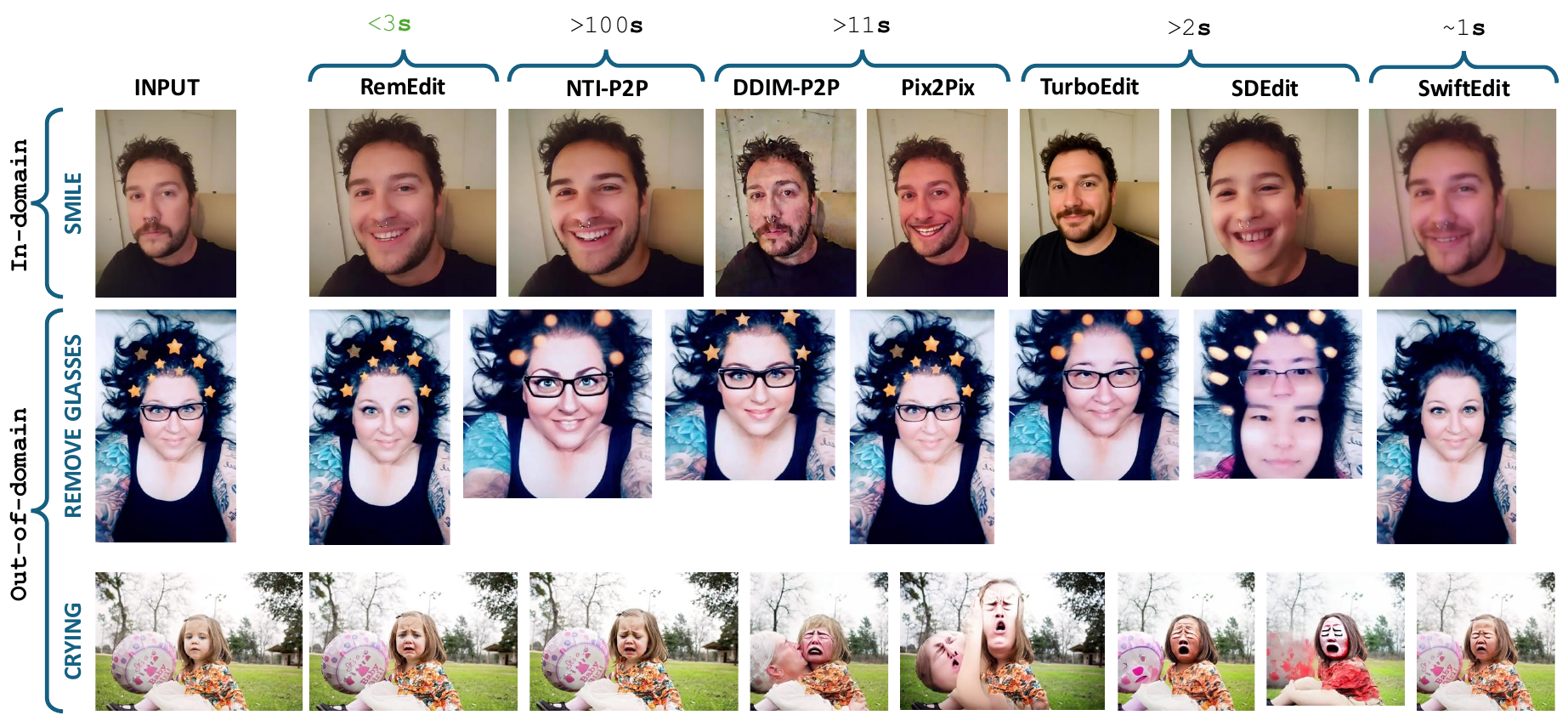}
    \caption{Zero shot qualitative comparison on RealEdit benchmark dataset. Identical prompts, inversion depth, and seeds. RemEdit achieves consistent attribute transfer with minimal collateral change while preserving identity.\vspace{-0.5cm}}
\end{figure}
%
% %--------------- Table: CelebA-HQ -----------------
\begin{table*}[t]
  \centering
  \small
  \caption{\label{tab:celeba}Quantitative comparison of diffusion-based image editing methods on CelebA-HQ dataset. Results show ($S_{dir}$), Seg. Cons., and inference time across three facial attribute editing tasks (Smiling, Sad, Tanned).\vspace{-0.25cm}}
  \resizebox{\textwidth}{!}{
  \begin{tabular}{l c c c c c c c}
    \hline
    & \multicolumn{2}{c}{\textbf{Smiling}} 
    & \multicolumn{2}{c}{\textbf{Sad}} 
    & \multicolumn{2}{c}{\textbf{Tanned}}
    & \textbf{Inference Time} \\ 
    \cline{2-3}\cline{4-5}\cline{6-7}
    Method               
      & $S_{dir}$ $\uparrow$ & Seg.\ Cons.\ (\%) $\uparrow$
      & $S_{dir}$ $\uparrow$ & Seg.\ Cons.\ (\%) $\uparrow$
      & $S_{dir}$ $\uparrow$ & Seg.\ Cons.\ (\%) $\uparrow$
      & (sec) $\downarrow$ \\ \hline
    StyleCLIP~\cite{patashnik2021styleclip}         
        & 0.130 & 86.80 & 0.149 & 85.50 & 0.152 & 84.30 & 8.5 \\
    StyleGAN-NADA~\cite{gal2022stylegan}
        & 0.160 & 89.40 & 0.161 & 87.70 & 0.166 & 88.50 & 12.3 \\
    Diffusion-CLIP~\cite{kim2022diffusionclip}
        & 0.170 & \textbf{93.70} & 0.163 & \textbf{89.93} & 0.174 & \textbf{92.85} & 45.2 \\
    BoundaryDiffusion~\cite{zhu2023boundary}
        & 0.170 & 90.40 & \underline{0.166} & 89.02 & \underline{0.177} & 85.71 & 38.7 \\
    Asyrp~\cite{kwon2022diffusion} 
        & \underline{0.190} & 87.90 & 0.159 & 88.90 & \underline{0.177} & 89.31 & 28.9 \\
    Prompt-to-prompt (p2p)~\cite{hertz2022prompt}
        & 0.165 & 85.20 & 0.152 & 84.10 & 0.158 & 86.30 & 145.0 \\
    LEdits++~\cite{brack2024ledits++}
        & 0.182 & 89.70 & 0.169 & 87.80 & 0.175 & 88.90 & 20.1 \\ \hline
    %-------------------------------------------------------------------------
    \rowcolor{gray!30}\textbf{RemEdit}   
        & \textbf{0.1982}  & \underline{92.41} & \textbf{0.1792} & \underline{89.72} & \textbf{0.1948} & \underline{92.18} & \textbf{2.8} \\ \hline
  \end{tabular}}
\end{table*}

\begin{table*}[t]
  \centering
  \small
  \caption{\label{tab:lsun}Quantitative evaluation on the \textbf{LSUN-Church} dataset.\vspace{-0.25cm}}
  \setlength{\tabcolsep}{6pt}
  \begin{tabular}{l c c c c c c}
    \hline
    & \multicolumn{2}{c}{\textbf{Department Store}} 
    & \multicolumn{2}{c}{\textbf{Ancient}} 
    & \multicolumn{2}{c}{\textbf{Red Brick}} \\ 
    \cline{2-3}\cline{4-5}\cline{6-7}
    Method
      & $S_{dir}$ $\uparrow$ & Seg.\ Cons.\ (\%) $\uparrow$
      & $S_{dir}$ $\uparrow$ & Seg.\ Cons.\ (\%) $\uparrow$
      & $S_{dir}$ $\uparrow$ & Seg.\ Cons.\ (\%) $\uparrow$ \\ \hline
    Diffusion CLIP~\cite{kim2022diffusionclip} 
        & 0.1300 & 54.50 & 0.1976 & \textbf{64.82} & 0.2085 & 65.83 \\
    BoundaryDiffusion~\cite{zhu2023boundary}
        & 0.1866 & 56.72 & 0.2034 & 60.13 & 0.2112 & 66.10 \\
    Asyrp~\cite{kwon2022diffusion}
        & 0.1932 & 57.62 & 0.2087 & 62.65 & 0.2170 & 67.42 \\
    Prompt-to-prompt (p2p)~\cite{hertz2022prompt}
        & 0.1820 & 53.40 & 0.1890 & 58.75 & 0.1985 & 61.30 \\
    LEdits++~\cite{brack2024ledits++}
        & 0.1895 & \underline{58.85} & 0.2045 & 59.20 & 0.2130 & 64.95 \\ \hline
    %-------------------------------------------------------------------------
    \rowcolor{gray!30}\textbf{RemEdit}
        & \textbf{0.1959} & \textbf{59.03} & \textbf{0.2193} & 63.91 & \textbf{0.2200} & \textbf{68.20} \\ \hline
  \end{tabular}
\end{table*}

\begin{table}[t]
\centering
\caption{\label{tab:reconstruction_quality}
Reconstruction fidelity of \emph{RemEdit} under different inversion depths $t_{0}$, forward DDIM steps $S_{\text{for}}$, and generative refinement steps $S_{\text{gen}}$. Each cell reports MAE / LPIPS / SSIM. Lower MAE/LPIPS and higher SSIM indicate better fidelity.\vspace{-0.2cm}}
\scriptsize
\resizebox{0.475\textwidth}{!}{
    \begin{tabular}{lccc}
    \hline
    & \multicolumn{3}{c}{$\boldsymbol{S_{\text{gen}}}$} \\
    \cline{2-4}
    & 6 & 40 & 500 \\
    \hline
    \rowcolor{gray!30}\multicolumn{4}{c}{$t_{0} = 300$} \\
    \hline
    $\boldsymbol{S_{\text{for}}} = 6$ 
      & 0.047 / 0.185 / 0.732
      & 0.061 / 0.221 / 0.704
      & 0.063 / 0.224 / 0.694 \\
    $\boldsymbol{S_{\text{for}}} = 40$ 
      & 0.027 / 0.110 / 0.863
      & 0.023 / 0.091 / 0.891
      & 0.023 / 0.086 / 0.895 \\
    $\boldsymbol{S_{\text{for}}} = 500$ 
      & 0.024 / 0.095 / 0.885
      & 0.020 / 0.073 / 0.914
      & 0.019 / 0.065 / 0.923 \\
    \hline
    \rowcolor{gray!30}\multicolumn{4}{c}{$t_{0} = 450$} \\
    \hline
    $\boldsymbol{S_{\text{for}}} = 6$ 
      & 0.055 / 0.208 / 0.673
      & 0.073 / 0.255 / 0.655
      & 0.077 / 0.260 / 0.643 \\
    $\boldsymbol{S_{\text{for}}} = 40$ 
      & 0.031 / 0.128 / 0.827
      & 0.025 / 0.100 / 0.880
      & 0.024 / 0.093 / 0.885 \\
    $\boldsymbol{S_{\text{for}}} = 500$ 
      & 0.028 / 0.108 / 0.862
      & 0.024 / 0.076 / 0.910
      & 0.020 / 0.068 / 0.919 \\
    \hline
    \rowcolor{gray!30}\multicolumn{4}{c}{$t_{0} = 600$} \\
    \hline
    $\boldsymbol{S_{\text{for}}} = 6$ 
      & 0.084 / 0.283 / 0.501
      & 0.101 / 0.325 / 0.564
      & 0.106 / 0.330 / 0.552 \\
    $\boldsymbol{S_{\text{for}}} = 40$ 
      & 0.047 / 0.175 / 0.706
      & 0.029 / 0.120 / 0.852
      & 0.028 / 0.108 / 0.862 \\
    $\boldsymbol{S_{\text{for}}} = 500$ 
      & 0.041 / 0.147 / 0.778
      & 0.024 / 0.087 / 0.893
      & 0.022 / 0.076 / 0.907 \\
    \hline
    \end{tabular}
    }
\vspace{-0.3cm}
\end{table}
\begin{table}[ht]
\centering
\caption{\label{tab:ablation}Ablation study on the core components of RemEdit %for the ``Smiling'' attribute on CelebA-HQ. 
The results show that each component provides a distinct benefit. \textbf{Time (s) is measured end-to-end, including VLM captioning, inversion, geodesic ODE solving, and decoding.}}
\resizebox{\columnwidth}{!}{%
    \begin{tabular}{@{}lccc@{}}
    \toprule
    \textbf{Method Configuration} & \textbf{Dir. CLIP} $\uparrow$ & \textbf{Seg. Cons. (\%)} $\uparrow$ & \textbf{Time (s)} $\downarrow$ \\
    \midrule
    \textbf{Euclidean offset ($h$-space baseline)} & 0.190 & 87.9 & {\bf 1.82} \\
    + Geodesic Navigation (Ours) & 0.196 & 89.5 & 2.76 \\
    \midrule
    Linear Interpolation (no SLERP) & 0.188 & 86.7 & 2.74 \\
    Inner-only SLERP & 0.191 & 88.2 & 2.78 \\
    Outer-only SLERP & 0.192 & 88.4 & 2.77 \\
    Dual-SLERP Blending (Ours, unpruned) & \textbf{0.198} & \textbf{92.4} & 2.89 \\
    \midrule
    Full RemEdit + Pruning ($\rho=0.2$) & 0.192 & 90.1 & 2.38 \\
    Full RemEdit + Pruning ($\rho=0.5$) & 0.184 & 89.5 & 2.31 \\
    \bottomrule
    \end{tabular}%
    }
    \vspace{-.2in}
\end{table}

% \begin{table}[ht]
% \centering
% \caption{\label{tab:ablation}Ablation study on the core components of RemEdit for the ``Smiling'' attribute on CelebA-HQ. The results show that each component provides a distinct benefit. \textcolor{blue}{\textbf{Time (s) is measured end-to-end, including the ODE integration overhead}}}
% \resizebox{\columnwidth}{!}{%
%     \begin{tabular}{@{}lcccc@{}}
%     \toprule
%     \textbf{Method Configuration} & \textbf{Dir. CLIP} $\uparrow$ & \textbf{Seg. Cons. (\%)} $\uparrow$ & \textbf{Time (s)} $\downarrow$ \\
%     \midrule
%     Baseline (Asyrp-style h-space edit) & 0.190 & 87.9 & {\bf 1.82} \\
%     + Geodesic Navigation (Ours) & 0.196 & 89.5 & 2.76 \\
%     \parbox{4cm}{
%     + Dual-SLERP Blending \\ (Full RemEdit, unpruned)} & \textbf{0.198} & \textbf{92.4} & 2.89 \\
%     \midrule
%     Full RemEdit + Pruning ($\rho=0.2$) & 0.192 & 90.1 & 2.38 \\
%     Full RemEdit + Pruning ($\rho=0.5$) & 0.184 & 89.5 & 2.31 \\
%     \bottomrule
%     \end{tabular}%
%     }
%     \vspace{-.2in}
% \end{table}

% \vspace{-0.5cm}
\section{Experiments}
\label{sec:experiments}

    We conduct a comprehensive set of experiments to validate \emph{RemEdit}: the accuracy, the effectiveness of our semantic control mechanisms, and the efficiency of our task-aware pruner. We evaluate our method on several challenging benchmarks, including CelebA-HQ~\cite{karras2017progressive}, LSUN-Church~\cite{journals/corr/YuZSSX15}, and AFHQ-Dog~\cite{Choi_et_al_2024} each over $256\times256$.

    \subsection{Implementation Details}
    Our framework is trained in an efficient few-shot manner. Unlike baseline methods such as Asyrp~\cite{kwon2022diffusion} which require multiple iterations over $1000+$ images, RemEdit achieves superior results by training on only $500$ images for two-shot setup ($n$=2). We observe that a single training iteration is sufficient for the Riemannian block to capture the high-level semantics of an edit, though training for approximately three iterations significantly refines the output by removing minor artifacts in detailed regions like eyebrows and hair texture. All experiments are conducted using the DDIM architecture on $256\times256$ resolution images unless otherwise specified. We report wall-clock end-to-end inference measurements for the pipeline.

\subsection{Analysis of Geometric Semantic Control}
\label{sec:exp-analysis}

    \subsubsection{Quantitative Analysis}
    We first evaluate RemEdit’s core editing, highlighting the synergy between geodesic navigation and dual-SLERP blending. Following recent SOTA methods~\cite{kwon2022diffusion,zhu2023boundary,kim2022diffusionclip}, Tab.~\ref{tab:celeba} and \ref{tab:lsun} compare RemEdit on CelebA-HQ and LSUN-Church using directional CLIP similarity ($S_{\text{dir}}$)~\cite{radford2021learningCLIP} for semantic alignment and segmentation consistency~\cite{lee2020maskganSC} for identity preservation. We chose 250 samples per task attribute. RemEdit consistently outperforms prior work, including Asyrp, across attributes like Smiling,'' Sad,’’ and ``Tanned.’’ This highlights the effectiveness of our exponential map: by solving a geodesic ODE, the Riemannian block ensures semantically precise edits, while dual-SLERP preserves identity.

    \subsubsection{Qualitative Analysis}
    Qualitative results across attributes are shown in Fig.~\ref{fig:various_attr}. RemEdit successfully manipulates attributes from simple expressions (``\texttt{Sad},'' ``\texttt{Smiling}'') to more complex transformations (``\texttt{Zombie},'' ``\texttt{Pixar}'') while maintaining a high degree of photorealism and identity. Fig.~\ref{fig:diffucult-case} provides a direct comparison against the Asyrp baseline. Even when running with only 40 denoising steps, RemEdit produces results that are qualitatively superior to Asyrp running at 1000 steps, successfully editing difficult cases where the baseline struggles. This highlights the efficiency and robustness of our geometric approach. As seen in Fig.~\ref{fig:qwen2}, Qwen2-VL improves clarity on underspecified prompts, whereas regular cases and the underlying geometry remain stable.

    \subsubsection{Ablation Study}
    \paragraph{Reconstruction Quality.}Tab.~\ref{tab:reconstruction_quality} investigates the reconstruction fidelity of our model under various inversion and generation settings. We analyze the impact of the inversion depth ($t_0$), the number of forward DDIM steps ($S_{for}$), and the number of generative steps ($S_{gen}$). The results, measured by MAE, LPIPS, and SSIM, show that a deeper inversion (higher ($t_0$) and a sufficient number of generative steps are crucial for high-fidelity reconstruction, confirming that our method adheres to the expected behavior of diffusion-based inversion pipelines.

        \vspace{-.1in}
    \paragraph{Controlled Image Generation.}We also conduct an ablation study on individual contributions, with results presented in Tab.~\ref{tab:ablation}. We start with a strong baseline implementing an Asyrp-style h-space edit. We then incrementally add our proposed modules: first the geodesic navigation, then the dual-SLERP blending to form the full, unpruned RemEdit model, and finally our task-specific pruning at two different ratios ($\rho=0.2$ and $\rho=0.5$). We evaluate each configuration on directional CLIP similarity $S_{dir}$, segmentation consistency $SC$, and inference time per image.
    
\subsection{Analysis of Task-Specific Acceleration}
\label{sec:acceleration}

\begin{figure}
    \centering
    \includegraphics[width=0.9\linewidth]{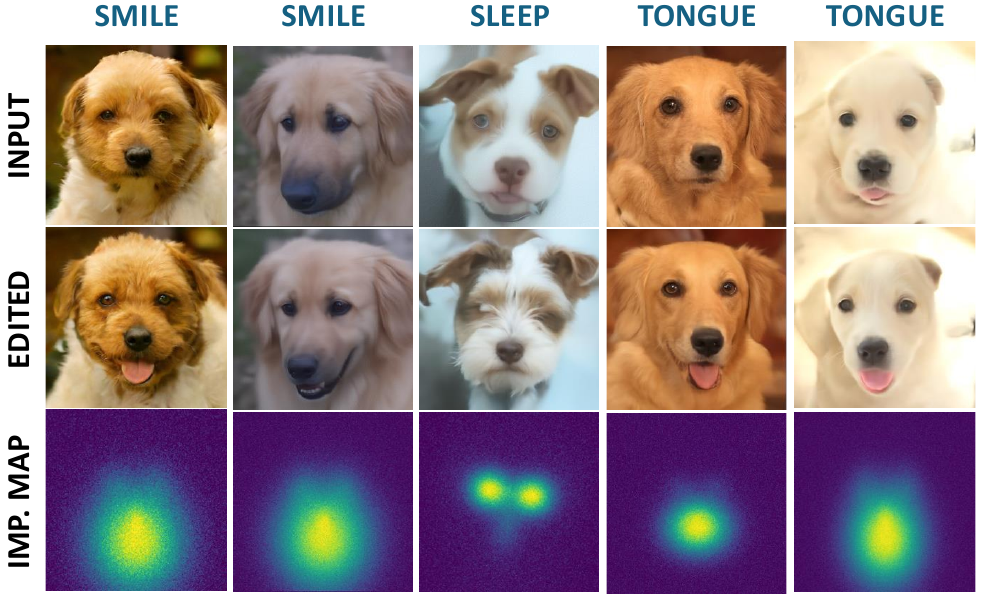}
    \caption{\label{fig:importance-map}Token importance map heatmap per image for visualizing what the pruning head attends to.}
    \vspace{-0.2cm}
\end{figure}

    %Qualitative Analysis. 
    As shown in Fig.~\ref{fig:teaser}, RemEdit maintains fidelity even under aggressive pruning. At a 10\% pruning ratio, the edited output is visually indistinguishable from the unpruned version, while still providing a ~10\% speed-up. Even at an aggressive 90\% pruning ratio, which yields an $\sim$\!18\% speed-up, the edit remains semantically correct and visually acceptable. This degradation is a direct result of our task-aware approach. Token retention patterns are illustrated in Fig.~\ref{fig:importance-map}. The importance maps clearly show that the model learns to focus on the semantically relevant regions (e.g., the mouth for ``\texttt{Smiling},'' the entire face for ``\texttt{Sleep}'') while correctly identifying the background as prunable.

\begin{figure}
    \centering
    \includegraphics[width=0.8\linewidth]{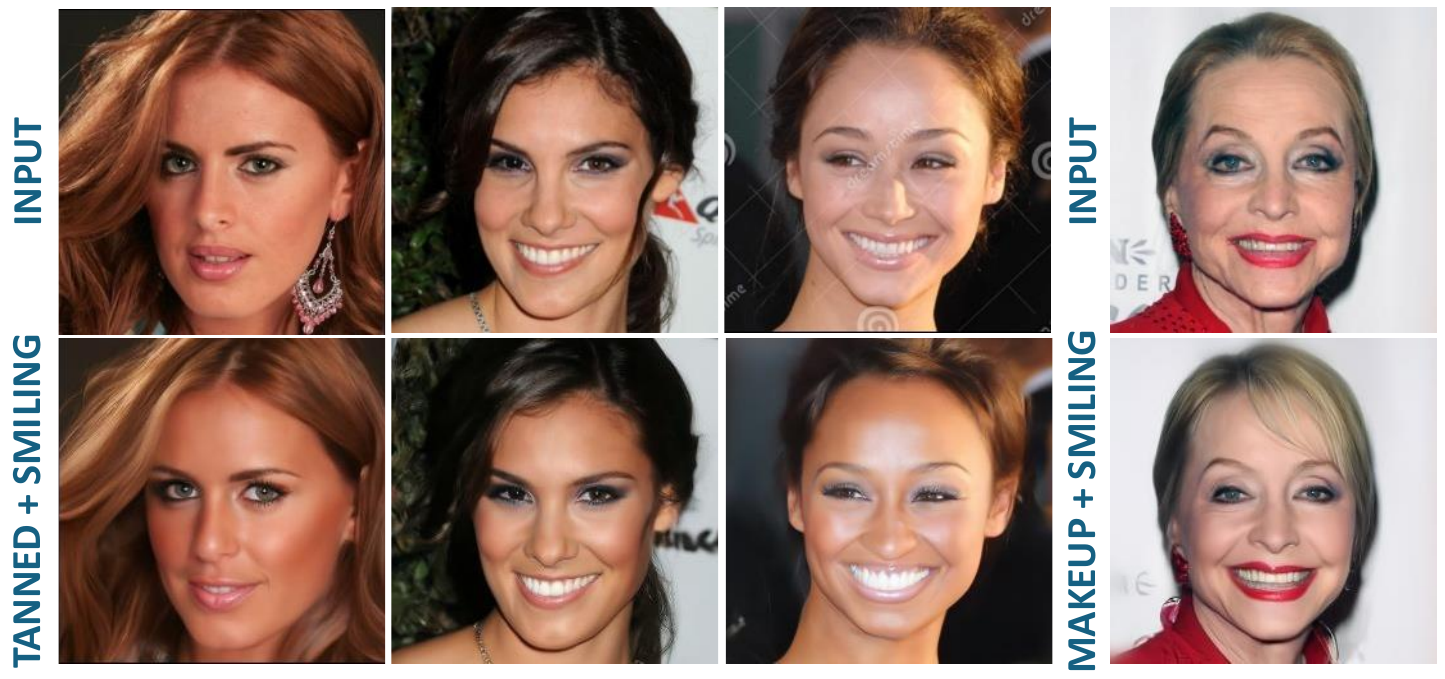}
    \caption{\label{fig:multi-attr}We also demonstrate the semantic control capability of RemEdit through multi-attribute editing.}
    \vspace{-0.7cm}
\end{figure}
\vspace{-.05in}
\section{Results and Discussion}
\label{sec:results}

    Our experimental results validate the effectiveness of the RemEdit framework. The quantitative comparisons in Tables~\ref{tab:celeba} and \ref{tab:lsun} demonstrate a clear improvement over existing methods in semantic alignment and identity preservation.

    The ablation study in Tab.~\ref{tab:ablation} provides a deeper insight into the trade-offs involved.\ Our contributions on the geodesic approach and dual-SLERP blending modules significantly improve both the $S_{dir}$ score (from $0.190$ to $0.198$) and segmentation consistency (from $87.9\%$ to $92.4\%$), confirming their role in enhancing edit quality and fidelity. This gain, however, comes at the cost of increased inference time, rising from 1.82s for the baseline to 2.89s for the full unpruned RemEdit, due to the ODE-solving computation.

    This is where our task-specific pruning demonstrates its value. By pruning just 20\% of the tokens, we 
    %recover a significant portion of the speed, reducing the
    reduce inference time to $2.38s$ while maintaining performance metrics close to the full model. Even with an aggressive $50\%$ pruning ratio, the model remains competitive with the baseline in quality ($S_{dir} = 0.184~\text{vs}~0.190$) while being substantially faster than the unpruned version.\ It confirms that our pruning method successfully makes our geometric editing framework more practical, closing the performance gap with simpler methods while retaining a clear advantage in edit quality.

\begin{figure}
    \centering
    \includegraphics[width=0.9\linewidth]{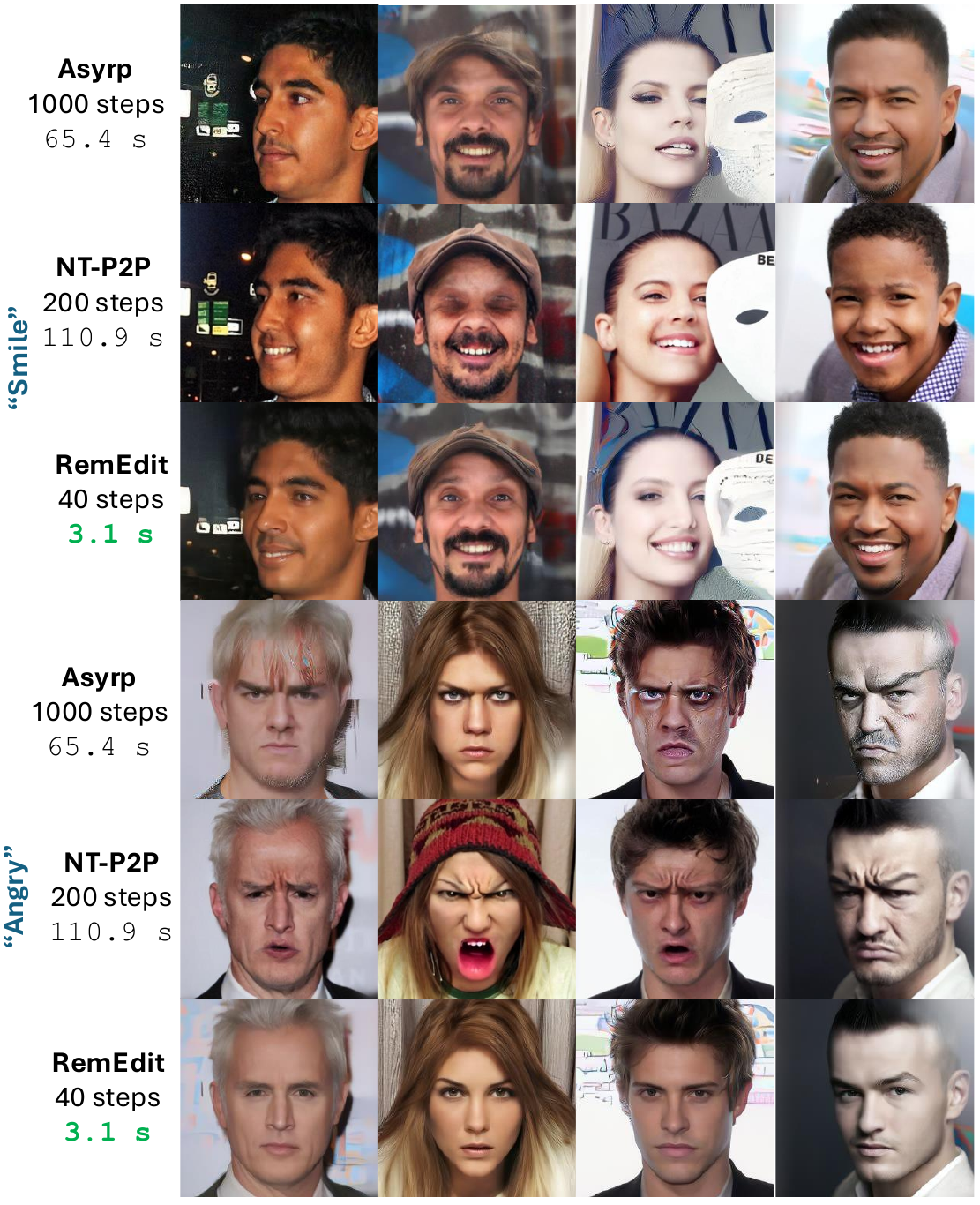}
    \caption{\label{fig:diffucult-case}Solving difficult cases in only 40 steps.}
    \vspace{-0.7cm}
\end{figure}

    % Fig.~\ref{fig:qwen2} On challenging, underspecified prompts, one optional Qwen2-VL pass yields clearer edits; regular cases appear similar with or without VLM; geometry is unchanged. 
    Furthermore, the token-importance maps in Fig.~\ref{fig:importance-map} confirm our task-aware pruning strategy by focusing on relevant regions (mouth, tongue, face) and assign lower background weights; quality is stable with 20–50\% pruning. Fig.~\ref{fig:multi-attr} shows multi-attribute examples (Tanned+Smiling, Makeup+Smiling) with no interference or identity change observed here. On difficult cases as in Fig.~\ref{fig:diffucult-case}, RemEdit reaches edit in \~{}40 steps (\~{}3.1s), Asyrp \~{}1000 (\~{}65.4s), NT-P2P \~{}200 (\~{}110.9s); panels show differences in identity and locality at the cost of inference speed.
    % Fig. 6. In the shown challenging cases with underspecified prompts, we observe that a single optional Qwen2-VL pass yields clearer edits, while regular cases look similar with or without VLM; the geometry modules are identical across settings.
    % Fig. 7. The token-importance maps concentrate on task-relevant regions (mouth for Smile, tongue for Tongue, face for Sleep) and show lower weights in background; in these examples, quality remains stable at 20–50\% pruning.
    % Fig. 8. The multi-attribute examples (Tanned+Smiling, Makeup+Smiling) display both attributes together, and in these cases we do not observe noticeable cross-attribute interference or identity change.    
    % Fig. 9. On the difficult cases shown, RemEdit reaches the target edit in about 40 steps (about 3.1 s), Asyrp uses about 1000 steps (about 65.4 s), and NT-P2P about 200 steps (about 110.9 s); the side-by-side panels show differences in identity similarity and edit locality.

\vspace{-.05in}
\section{Conclusion}
\label{sec:conclusion}
    In this work, we introduced RemEdit, a novel framework for high-fidelity and efficient controlled image generation. We addressed the fundamental trade-off between accuracy and speed by making contributions on three fronts. First, we proposed a new way to navigate the semantic $h$-space by modeling it as a Riemannian manifold and solving for geodesic paths, which we showed improves semantic quality. Second, we introduced a dual-SLERP and Qwen2-VL based text-injection blending mechanism for fine-grained control via textual and image features over edit strength and identity preservation, further boosting fidelity. Finally, we developed a task-aware attention-pruning method that makes our geometrically sophisticated approach computationally practical, significantly accelerating inference while preserving the quality of the edit. 

    While RemEdit demonstrates high semantic fidelity and practical efficiency, it relies on pretrained models such as Qwen2-VL and CLIP. Future work could explore learned, domain-specific guidance for out-of-distribution tasks.
    
{
    \small
    \bibliographystyle{ieeenat_fullname}
    \bibliography{main}
}

\clearpage\newpage \newpage
\section{Appendix}
\label{sec:supp}

\section*{A. Problem Statement}
\label{supp:extended_problem_statement}
    The goal of diffusion-based image editing is to develop a framework that is simultaneously faithful to user intent, geometrically accurate, and computationally efficient. While prior work has shown that adding a semantic offset $\Delta{h}$ to the bottleneck features in $h$-space is an effective manipulation strategy~\cite{kwon2022diffusion,nguyen2025hedit}, this approach presents a tripartite challenge that has not been holistically addressed.

    First, the standard operation, $h' = h + \Delta{h}$, implicitly treats this space as Euclidean, which can lead to artifacts and off-manifold results. This leaves a foundational question unresolved:
    \begin{center}
        \textbf{What is the optimal formulation for the edit vector $\Delta{h}$ that respects the intrinsic, non-Euclidean structure of the data manifold?}
    \end{center}
    
    Second, even with a geometrically sound direction, applying the edit requires fine-grained control to preserve the subject's identity and avoid artifacts. This is made more difficult by the inherent ambiguity of simple text prompts, which can lead to unintended semantic shifts. This raises the question:
    \begin{center}
        \textbf{How can an edit be applied with tunable strength and guided by precise, context-aware instructions to ensure high fidelity?}
    \end{center}
    
    Finally, any method that increases model sophistication to improve fidelity risks exacerbating the already significant computational cost of diffusion models, hindering practical application. This leads to the third challenge:
    \begin{center}
        \textbf{How can a high-fidelity editing process be made computationally efficient without compromising the quality and semantic consistency of the edit?}
    \end{center}
    
    This paper addresses these interconnected problems. We hypothesize that a truly robust solution requires a unified framework: one that models $h$-space as a Riemannian manifold to derive a principled edit vector, introduces advanced blending and guidance mechanisms for high-fidelity control, and integrates a task-aware acceleration strategy to ensure efficiency. Our work aims to formalize and implement such a holistic framework.

\section*{B. Details on Task-aware Pruning}
\label{supp:extended_details_prune}
    \begin{enumerate}
        \item \textbf{Importance Scoring:}
        The input $X$ is reshaped into a token sequence $T \in \mathbb{R}^{B \times N \times C}$, where $N=H \times W$. The \texttt{PruningHead} function $\mathcal{P}_\theta$ computes importance scores $S$:
        \begin{equation}
            S = \mathcal{P}_\theta(T, \mathbf{d}_{\text{edit}}) \in [0, 1]^{B \times N}
        \end{equation}

        \item \textbf{Index Selection:}
        Given a pruning ratio $\rho$, we keep $k = \lfloor N \cdot (1 - \rho) \rfloor$ tokens. The indices $\mathcal{I}_{\text{keep}}$ of these tokens are selected:
        \begin{equation}
            \mathcal{I}_{\text{keep}} = \text{topk}_{\text{indices}}(S, k)
        \end{equation}
    
        \item \textbf{Pruned Attention:}
        The query ($Q$), key ($K$), and value ($V$) projections are gathered using the selected indices to form pruned sets $Q_{\text{kept}}, K_{\text{kept}}, V_{\text{kept}} \in \mathbb{R}^{B \times k \times C}$. Attention is computed only on this reduced set:
        \begin{equation}
            A_{\text{pruned}} = \text{Softmax}\left(\frac{Q_{\text{kept}} K_{\text{kept}}^T}{\sqrt{C}}\right) V_{\text{kept}}
        \end{equation}
    
        \item \textbf{Scattering and Output:}
        The attended features $A_{\text{pruned}} \in \mathbb{R}^{B \times k \times C}$ are scattered back into a zero tensor of the original size, $A_{\text{result}} \in \mathbb{R}^{B \times N \times C}$, at their original positions $\mathcal{I}_{\text{keep}}$. The final output $X_{\text{out}}$ is computed via the residual connection with the output projection $W_o$:
        \begin{equation}
            X_{\text{out}} = X + W_{o}(A_{\text{result}})
        \end{equation}
    \end{enumerate}

\section*{C. Exponential Map Architecture}
\label{fig:expomap}

    \begin{figure}[ht]
        \centering
        \includegraphics[width=\linewidth]{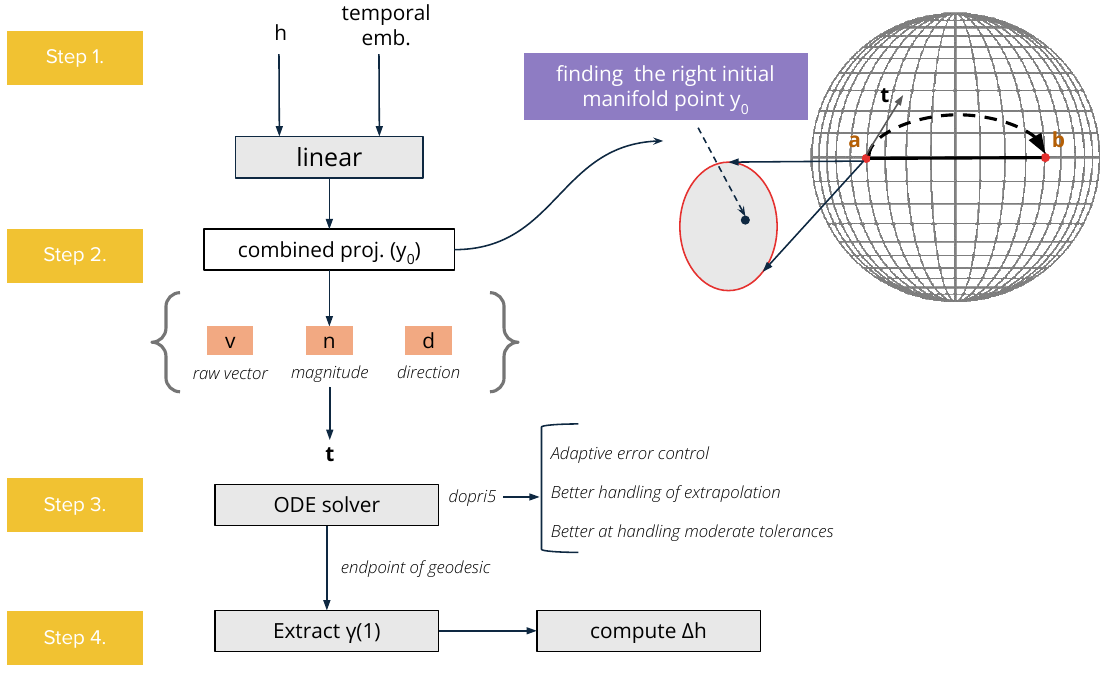}
        \caption{\label{fig:expomap_arch}Our ExpoMap module learns the local geometry of the $h$-space manifold by estimating Christoffel symbols and solving the geodesic ODE using a numerical integrator (\texttt{Dopri5}). It takes as input a semantic direction and timestep, maps it through a learnable projection to obtain the initial tangent vector $v_{0}$, and integrates along the manifold to compute the final offset $\Delta{h} = \gamma(1)-h$. This enables geometry-aware edits without post-hoc Jacobian estimation.}
    \end{figure}

\begin{table*}[!ht]
\centering
\caption{\label{tab:lerp_vs_slerp}Comparison of interpolation methods for diffusion model editing. While LERP is simple, it often produces artifacts by deviating from the data manifold. SLERP improves upon this by preserving the hyperspherical structure, and NoiseDiffusion further refines this for natural images by correcting the noise distribution. Our proposed dual-SLERP provides granular control by operating on both the learned feature manifold and the noise space, enabling disentangled control over edit strength and identity preservation.}
\scalebox{0.63}{
    \begin{tabular}{@{}lllll@{}}
    \toprule
    \textbf{Feature} & \textbf{LERP (Linear)} & \textbf{SLERP (Spherical)}~\cite{shoemake1985animating,song2020denoising} & \textbf{NoiseDiffusion}~\cite{zheng2024noisediffusion} & \textbf{Dual-SLERP (Ours)} \\
    \midrule
    \textbf{Manifold Consistency} & Off-manifold & On hypersphere & Corrective (projects to valid noise distribution) & \textbf{Dual-manifold aware} (Riemannian \& Hyperspherical) \\
    \textbf{Semantic Blending} & Limited / Entangled & Fine-grained & Fine-grained (Corrected) & \textbf{Hierarchical \& Disentangled} \\
    \textbf{Artifact Risk} & High & Moderate (on natural images) & Lower (for natural images) & \textbf{Lowest} (due to orthogonality) \\
    \textbf{Control Mechanism} & Coarse (single parameter) & Smooth \& Structured & Multi-parameter (correction + interpolation) & \textbf{Disentangled control} (edit strength vs. fidelity) \\
    \bottomrule
    \end{tabular}%
}
\end{table*}

\section*{D. Workflow details on Qwen2-VL}
    While CLIP embeddings have proven effective for guiding semantic edits, they often encode social or aesthetic biases and operate within a relatively narrow distribution of concepts. To mitigate this limitation, we enrich the CLIP embedding with additional context from a frozen Qwen2-VL model, effectively forming a broader \texttt{CLIP}+\texttt{Qwen2} joint embedding space. This expanded representation captures fine-grained, instance-specific visual semantics—especially useful for under-specified prompts like ``face'' or ``young woman.'' While our Riemannian editing and dual-SLERP modules refine the geometry of noise and feature spaces, this multimodal enrichment ensures alignment at the textual level. Together, they allow RemEdit to perform edits that are both geometrically coherent and semantically precise, even under vague or biased attribute conditions.
    
    \begin{figure}[!ht]
        \centering
        \includegraphics[width=\linewidth]{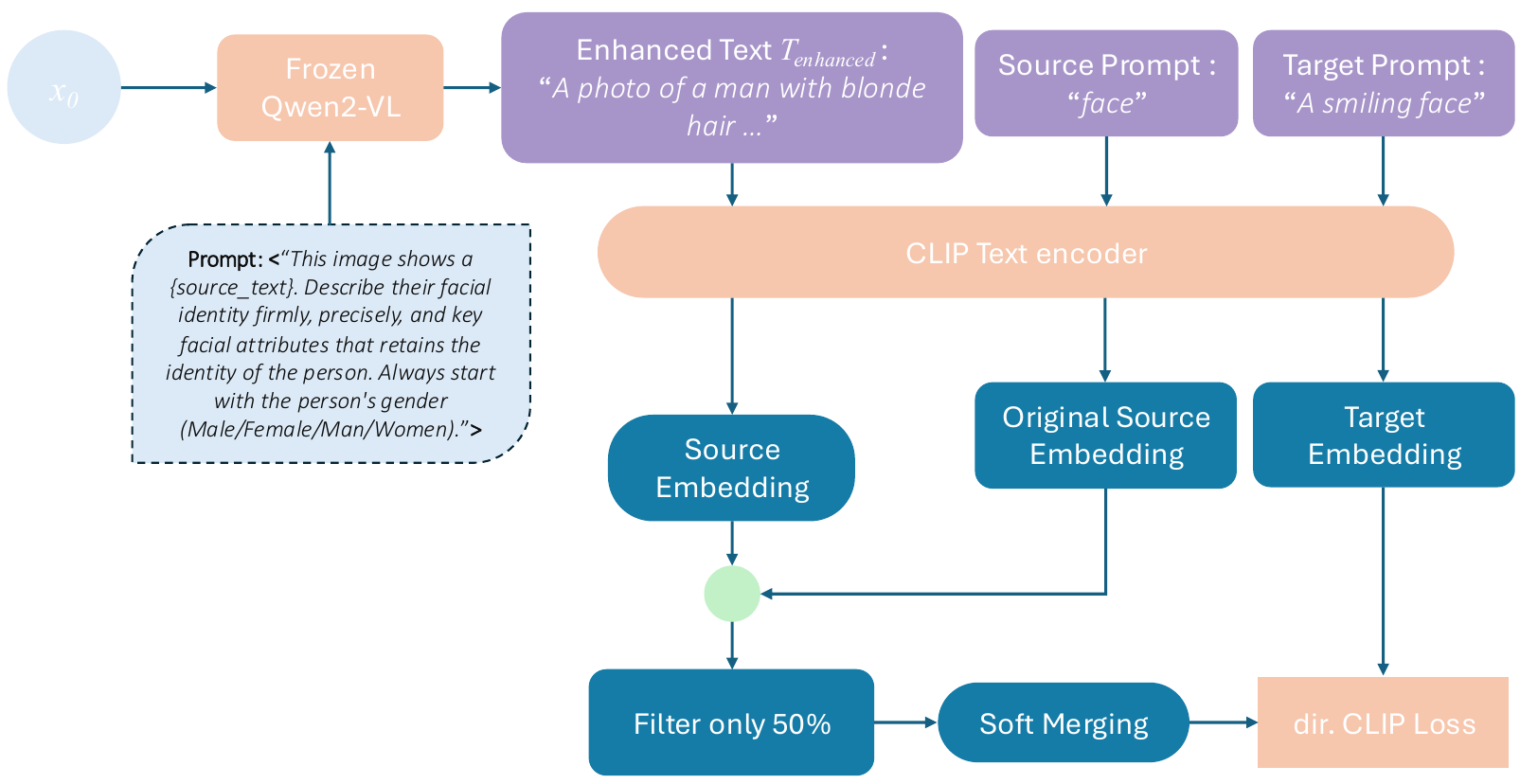}
        \caption{\label{fig:qwen-diagram}The source image $x_0$ and a descriptive prompt are fed into a frozen Qwen2-VL model to generate a detailed, instance-specific caption, $T_{\text{enhanced}}$. This enhanced text provides a richer source embedding, constraining the edit to be more faithful to the original image's core attributes.}
    \end{figure}

\section*{E. Image Interpolation in Diffusion Models}
    To better situate our contribution, we provide a conceptual comparison of different interpolation strategies in Tab.~\ref{tab:lerp_vs_slerp}. %The most basic method, Linear Interpolation (LERP), often fails in high-dimensional latent spaces as it does not respect the data manifold, leading to a high risk of artifacts and entangled semantic blending. Spherical Linear Interpolation (SLERP) provides a significant improvement by operating on the hypersphere, ensuring norm-preservation and offering smoother, more structured control. However, as noted by recent work, standard SLERP can still produce artifacts when interpolating real-world, natural images whose encoded noise deviates from the expected distribution. 
    NoiseDiffusion~\cite{zheng2024noisediffusion} mitigates this by projecting noise back to a valid prior. Our proposed dual-SLERP mechanism goes further, being dual-manifold aware: inner SLERP operates in the learned Riemannian $h$-space, while outer SLERP modulates fidelity in noise space. This disentangled scheme allows separate control over semantic strength and identity preservation, enabling precise and robust image editing.
% \clearpage\newpage \input{rebuttal}

\end{document}